%% file: root.tex
\documentclass[letterpaper, 10 pt, conference]{deps/ieeeconf}

\IEEEoverridecommandlockouts
\input{deps/arena}

\input{content/_defines}

\usepackage{listings}
\begin{document}

\maketitle
\thispagestyle{empty}
\pagestyle{empty}

\begin{abstract}
\input{content/0-abstract}
\end{abstract}

\begin{table*}[t]
\renewcommand{\arraystretch}{0.9}
\centering
\begin{tabularx}{\textwidth}{l *{5}{>{\centering\arraybackslash}X}}
\toprule
Method
  & Social Force
  & Navigation
  & Behavior
  & Heterogeneous
  & Natural Language
\\
\midrule
\multicolumn{6}{l}{\textit{SFM-based}} \\
PedsimROS
  & \cmark & \xmark & \xmark & \pmark & \xmark \\
MengeROS
  & \cmark & \xmark & \xmark & \pmark & \xmark \\
\midrule
\multicolumn{6}{l}{\textit{Learning-based}} \\
SocNavBench~\cite{socnavbench}
  & \xmark & \xmark & \xmark & \pmark & \xmark \\
SocialGym~2.0~\cite{socialgym}
  & \cmark & \xmark & \xmark & \pmark & \xmark \\
\midrule
\multicolumn{6}{l}{\textit{Scenario-based}} \\
SEAN~2.0~\cite{sean}
  & \xmark & \pmark & \cmark & \pmark & \xmark \\
Arena~5.0~\cite{arena:5}
  & \cmark & \xmark & \pmark & \cmark & \xmark \\
HuNavSim~2.0~\cite{hunavsim2}
  & \cmark & \xmark & \cmark & \cmark & \xmark \\
\midrule
\multicolumn{6}{l}{\textit{Other}} \\
Text-Crowd~\cite{textcrowd}
  & \xmark & \cmark & \xmark & \xmark & \cmark \\
TRACE~\&~PACE~\cite{tracepace}
  & \xmark & \cmark & \xmark & \xmark & \xmark \\
\midrule
\textbf{\WORKNAME~(Ours)}
  & \cmark & \cmark & \cmark & \cmark & \cmark \\
\bottomrule
\end{tabularx}
\caption{Feature comparison of popular pedestrian simulators.
\cmark~supports, \xmark~does not support, \pmark~partially supports.}
\label{tab:comparison}
\end{table*}

\section{Introduction}
\input{content/1-introduction}

\input{content/2-related}

\section{Methodology}

\input{content/3-methodology}

\section{Evaluations}

\input{content/4-evaluation}

\section{Conclusion}
\input{content/5-conclusion}


\bibliography{references.bib}

\end{document}

%% file: deps/arena.tex
\usepackage[utf8]{inputenc}
\usepackage[T1]{fontenc}
\usepackage{csquotes}
\usepackage{float}
\usepackage{amsmath,amssymb}
\usepackage{xcolor}
\usepackage{algpseudocodex}
\usepackage{array,tabularx,makecell,booktabs}
\usepackage{pifont}
\usepackage{rotating}
\usepackage{graphicx}
\usepackage{subcaption}
\usepackage{minted}  

\usepackage[bookmarks=false]{hyperref}
\hypersetup{
    colorlinks=true,
    linktoc=all,
    citecolor=black,
    filecolor=black,
    linkcolor=black,
    urlcolor=black
}

\usepackage{caption}
\usepackage{xspace}

\makeatletter
\DeclareRobustCommand\TODO{\textbf{TODO}\@latex@warning{TODO used}\xspace}
\makeatother

\newcommand{\aand}[2][]{#2$^{#1}$,}
\newcommand{\aandfinal}[2][]{#2$^{#1}$\vspace{0.2cm}\\}
\newcommand{\aattrib}[2][]{$^{#1}$#2\\}

\newcommand{\acaption}[2]{\caption[#1]{\emph{#1}:~#2}}

\newcommand{\cmark}{\textcolor{green!60!black}{\ding{51}}}
\newcommand{\xmark}{\textcolor{red!70!black}{\ding{55}}}
\newcommand{\pmark}{\textcolor{orange!80!black}{$\sim$}}

%% file: content/_defines.tex
\def\WORKNAME{GROVE\xspace}

\title{\LARGE \bf
\WORKNAME: Grounded Pedestrian Simulation via Natural Language for Interactive Social Robot Navigation
}

\author{
\aand[1]{Duc Tai Nguyen}
\aand[1, 2]{Volodymyr Shcherbyna}
\aand[1]{Anh Do Duc}\\
\aand[3]{Zhengcheng Shen}
\aand[3]{Teham Buiyan}
and~
\aandfinal[1]{Linh Kästner}
\aattrib[1]{Singapore Management University}
\aattrib[2]{Technical University Berlin}
\aattrib[3]{National University of Singapore}
}

\usepackage{multirow}

\usepackage{capt-of}

\makeatletter
\let\@oldmaketitle\@maketitle
\renewcommand{\@maketitle}{\@oldmaketitle
  \begin{center}
    \includegraphics[width=0.99\textwidth]{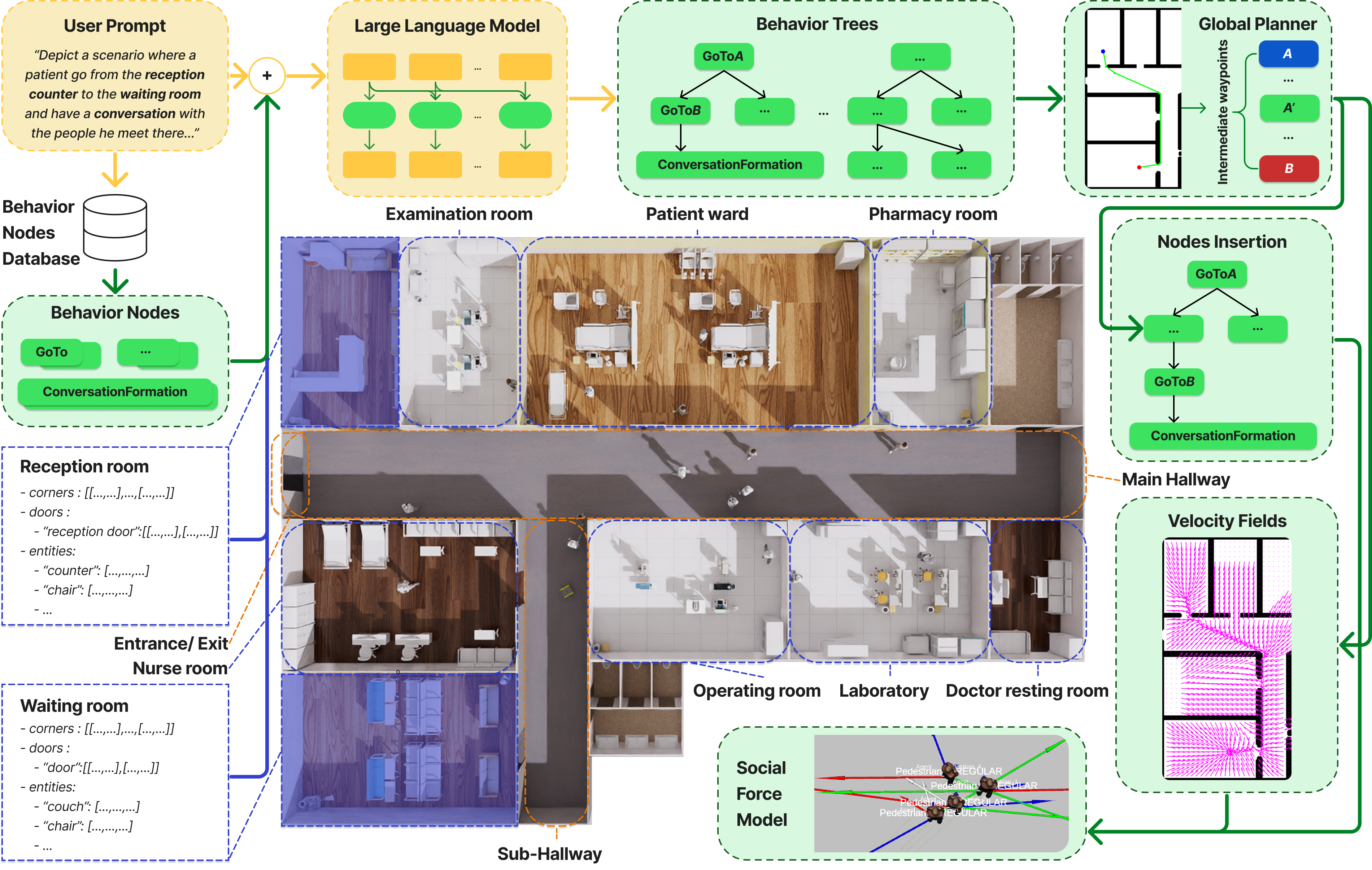}
    \captionof{figure}{\WORKNAME is a novel pedestrian scenario generation approach for social robotic simulation. \WORKNAME transforms natural language prompts into high-fidelity social simulations. Starting from a user prompt and world database, the pipeline extracts semantic Regions of Interest and utilizes Retrieval-Augmented Generation to fetch relevant Behavior Nodes. A Large Language Model synthesizes these inputs into behavior trees.  A Global Planner resolves navigation waypoints, which are either re-inserted into the behavior trees or converted into velocity fields based on the user preset. Finally, the post-processed behavior trees and a Social Force Model execute realistic human navigation and social interactions within simulators such as Isaac Sim or Gazebo.}
    \label{fig:intro}
  \end{center}
  \vspace{-10pt}
}
\makeatother

\usepackage{dblfloatfix}

%% file: content/0-abstract.tex
Pedestrian simulation is a critical component for training and deploying social robot navigation approaches, yet it remains a largely rigid system that repeatedly requires manual data generation to define even simple scenarios.
We propose \emph{\WORKNAME}, a text-to-scenario pedestrian simulation framework that combines state-of-the-art approaches to produce realistic, socially challenging scenarios for social robot navigation.
Our framework allows users to customize one of several common presets (emergency, queuing, normal) or even enter a fully independent prompt to generate a highly customizable pedestrian simulation.
Multiple modules separately ensure the realism and soundness of long-horizon human behavior, medium-horizon pedestrian navigation, and short-horizon robot/social interactions.
Each module is tuned by the prompt in a way that reflects the user intent across all aspects of pedestrian simulation.
By dynamically selecting one of several state-of-the-art (SotA) approaches in our modules based on the scenario, we capture many situational nuances of pedestrian behavior in order to narrow the simulation-to-real (sim2real) gap.
The human simulation is directly integrated into Isaac Sim, Gazebo, and RViz simulators for robot deployment in highly social environments.
We validate our approach through qualitative comparison against existing pedestrian simulation baselines across scenarios of varying complexity in residential, hospital, and office environments.
The result is a high-fidelity pedestrian simulation that challenges social robot navigation with complex, diverse, realistic human behaviors.

%% file: content/1-introduction.tex
Social robot navigation in human-populated environments remains one of the most challenging problems in mobile robotics.
Robots not only must plan collision-free paths, but also anticipate and respond to the complex, context-dependent behaviors of surrounding pedestrians~\cite{rudenko2020survey}.

Current pedestrian simulators share a common failure mode: they decouple scenario definition from the simulation itself, requiring researchers to manually specify agent positions, goals, and behaviors for every scenario they wish to test.
This manual bottleneck forces practitioners to work with small, fixed sets of hand-crafted scenarios~\cite{sean,arena:5} that fail to cover the long tail of real-world social situations.
Beyond this, most simulators apply uniform behavioral models across all agents~\cite{mavrogiannis2023eval}, collapsing the rich diversity of human pedestrian behavior into a single parameterization~\cite{chen2018sfmreview}.
They further rely on the Social Force Model (SFM)~\cite{helbing1995sfm, moussaid2010walking, chen2024social}, as the sole navigation primitive, a local collision avoidance mechanism never designed to guide agents through complex environments with semantic intent.

The result is a persistent sim-to-real gap~\cite{zhao2020sim2real}: simulated pedestrians that behave implausibly the moment they must navigate purposefully through a crowded scene.


We propose \WORKNAME, a generative framework combining RAG-based retrieval and social behavior trees for pedestrian data synthesis, closing this gap.
Given a free-text description, our framework automatically generates a fully configured, high-fidelity pedestrian simulation ready for robot deployment.

%% file: content/2-related.tex
\setcounter{figure}{1}
\begin{figure*}[t]
    \includegraphics[width=.99\linewidth]{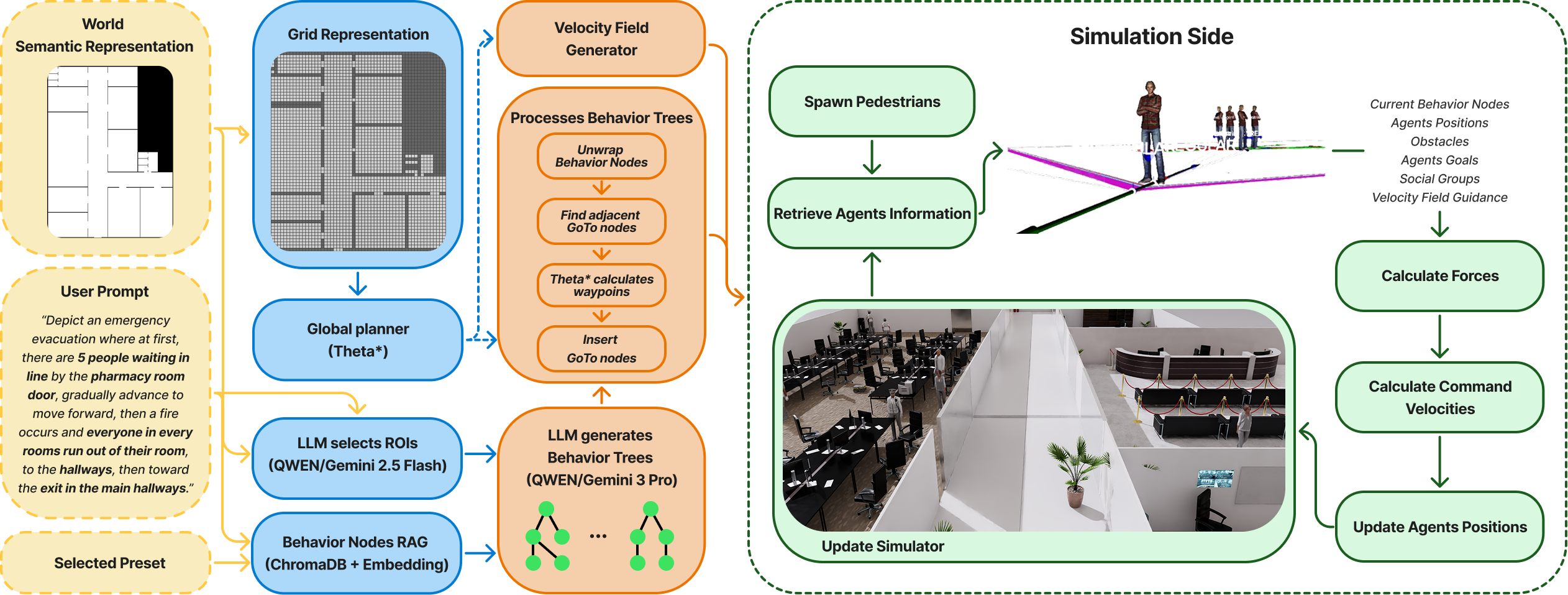}
    \acaption{System Design}{The framework employs a decoupled hierarchical approach to scenario generation. To mitigate model hallucination of complex behavioral logic, a RAG module fetches precise Behavior Node descriptions from a ChromaDB vector database. The LLM architecture utilizes a task-specific multi-model strategy: a lightweight model (Gemini 2.5 Flash~\cite{comanici2025gemini} or Qwen-3-0.6B~\cite{yang2025qwen3} on a local machine) performs rapid RoI selection, while a better reasoning model (Gemini 3 Pro or Qwen-3-4B) synthesizes error-free, semantically structured BTs. Navigation is resolved via a Theta* acting directly on the BTs to prevent local minima. Depending on the selected preset (dashed lines), waypoints are either injected as reactive nodes or transformed into velocity fields. The final execution loop integrates a Social Force Model to provide real-time, physically-consistent commands to the simulator.}
    \label{img:systemdesign}
\end{figure*}

\section{Related Work}
Numerous works study pedestrian representation and trajectory prediction~\cite{bokkasam2025pedestrian}, but most adopt a robot-centric or first-person perspective~\cite{li2025learning, jiang2025physgcn, shenkut2025visual}, limiting applicability to global, multi-agent crowd simulation.

Learning-based crowd generation methods fall into three directions.
Data-driven approaches learn pedestrian dynamics from real-world videos using diffusion or VAE-based models~\cite{yao2020learning, liu2024learning, bae2025continuous}, producing realistic motion but relying heavily on large-scale labeled data.
Physics-informed methods integrate diffusion models with social-force or fluid-based formulations~\cite{chen2024social, li2025efficient}, though at higher computational cost and with restricted behavioral diversity.
Text-guided approaches leverage LLMs to synthesize crowd behaviors from user prompts~\cite{textcrowd, shi2023learning, wang2023gensim}, improving flexibility and controllability, but remain largely disconnected from interactive robotics simulators.

On the robotics simulator side, PedsimROS and MengeROS~\cite{helbing1995sfm} brought SFM-based simulation into ROS but were deprecated alongside ROS~1 and assign uniform behavioral parameters to all agents.
SocialGym~2.0~\cite{socialgym} supports RL-based multi-robot training in structured social bottlenecks but inherits the same limitations.
SocNavBench~\cite{socnavbench} and nuScenes~\cite{caesar2020nuscenes} provide trajectory-level realism by replaying real pedestrians, but static trajectories prevent robot reactivity.
Arena~5.0~\cite{arena:5} offers Isaac Sim rendering, ROS~2 support, and modular scenario definitions, yet scenario generation remains template-driven.
SEAN~2.0~\cite{sean} introduces Behavior Graphs for group-level interactions but is restricted to Unity and ROS~1 with no mechanism for novel scenario generation.

HuNavSim~2.0~\cite{hunavsim2} achieves individual-level fidelity through per-agent behavior trees and supports Isaac Sim, Gazebo, and Webots.
Its core limitation is that behavior trees dispatch semantic intent to SFM without intermediate waypoints, producing implausible straight-line paths through complex environments, with all scenario configuration remaining manual.
We build directly on HuNavSim~2.0 for its powerful behavior tree engine.
Text-Crowd~\cite{textcrowd} demonstrates LLM-driven crowd generation from free-text but lacks simulator integration and interactive pedestrian-robot dynamics.
TRACE~\&~PACE~\cite{tracepace} generates realistic, waypoint-conditioned trajectories and, though not designed for robotics, directly addresses HuNavSim~2.0's navigation gap; we leverage it as an optional high-fidelity medium-horizon planner.

Our work addresses the gaps across all of the above: automatic text-driven scenario generation over HuNavSim~2.0, and Theta*-based waypoint planning~\cite{daniel2010theta} that bridges behavioral intent and physical navigation, with TRACE as a high-fidelity alternative, across long-, medium-, and short-horizon behavior.

%% file: content/3-methodology.tex
\subsection{System Design}

{
We propose a hierarchical crowd generation pipeline that separates semantic reasoning, behavioral planning, and geometric motion planning.
The system is divided into a generation side, responsible for semantic and behavioral reasoning, and a simulation side, responsible for geometric planning and execution, connected through a structured behavior tree interface.
To ensure scalability, robustness, and modularity, the architecture follows a hierarchical decomposition into three horizons.

\paragraph{Strategic layer} Regions of Interest (RoIs) are extracted from natural language input by matching the user prompt against candidates defined in a semantic world representation using an LLM.
\paragraph{Tactical layer} A structured database of behavior nodes is embedded into a vector store, and relevant nodes are retrieved via a Retrieval-Augmented Generation (RAG) module implemented with ChromaDB before being provided to the LLM for behavior tree generation that defines geometric goals and execution logic.
\paragraph{Operational layer} A global planning algorithm operates directly on navigation-related nodes in the generated behavior tree to compute static-obstacle collision-free paths for human agents.

\begin{figure*}[t]
    \centering
    \includegraphics[width=.99\linewidth]{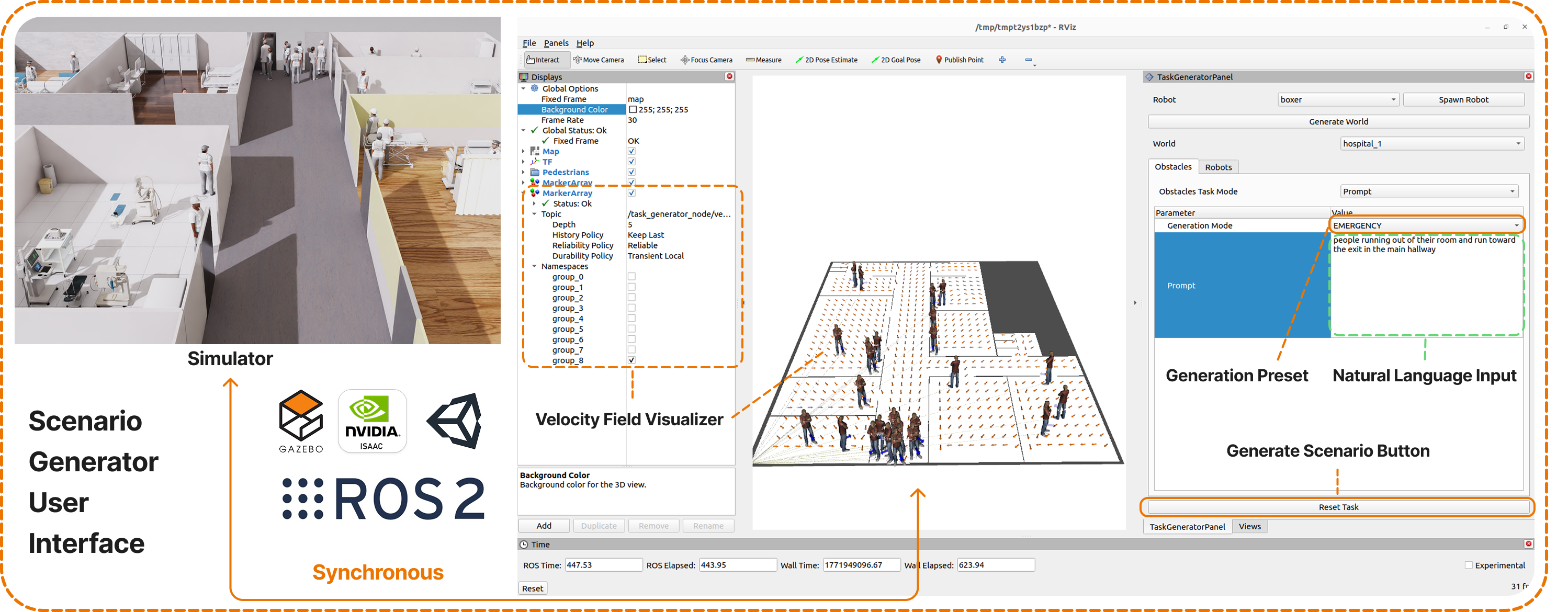}
    \acaption{User Story}{The proposed pipeline is integrated into the RViz2 visualization environment for enhanced usability and reproducibility. The UI allows users to input natural language prompts and select generation presets, which are then processed into visible artifacts. The interface provides a real-time synchronous link between the generation-side data—including velocity fields, global waypoints, and agent poses—and high-fidelity physics simulators such as Isaac Sim, Gazebo, and Unity via the ROS 2 ecosystem.}
    \label{img:ui}
\end{figure*}

\subsubsection{World Semantic Representation and RoIs extraction}

To bridge the gap between natural language intent and simulation geometry, we define a human-LLM-readable YAML format \texttt{world.yaml} that encodes spatial and semantic information in a structured and explicit manner.
The world is represented as a list of zones corresponding to RoIs such as rooms, sections, and hallways, each containing structural elements, entities, and semantic attributes enabling unambiguous reference from natural language.
An excerpt of \texttt{world.yaml} is shown in the left top panel of \autoref{fig:intro}.


RoIs are indexed at load time and stored with structured metadata consisting of zone name, door name, entity name, entity type, and associated geometric attributes.
During prompt processing, semantic tokens extracted from the user input are matched against this metadata to identify relevant RoIs.
Irrelevant zones are filtered out before constructing the LLM context.
For each selected RoI, a structured summary is constructed containing metadata, boundary coordinates, entrance coordinates, and selected entity information from \texttt{world.yaml}.
This filtered representation is injected into the LLM prompt to provide spatial grounding while avoiding unnecessary context expansion.


\subsubsection{LLM-Based Behavior Tree Generation with RAG}
\paragraph{Behavior Tree as Intermediate Representation}
behavior trees (BTs) are adopted as the intermediate representation due to their hierarchical structure, modularity, and deterministic execution semantics.
Unlike direct policy generation or trajectory synthesis, BTs provide an interpretable and verifiable symbolic program that mediates between language intent and simulation control.
This choice enables structured decomposition of complex crowd scenarios into reusable sub-behaviors while preserving execution transparency.
In our pipeline, the BT functions as a contract between the generation side and the simulation side, ensuring that all language-derived decisions are grounded in executable logic.

\paragraph{Node Wrapping Strategy}
Primitive HuNavSim nodes are insufficiently abstracted for language-driven generation because they expose low-level bookkeeping requirements such as explicit ID references for shared entities like group centers, leader agents, or navigation goals.
In complex scenarios with a large number of agents and navigation goals, maintaining consistent ID bindings becomes challenging for the LLM and leads to brittle generations.
We therefore introduce an automatic ID management module that deterministically allocates and resolves identifiers for agents, goals, and groups as a post-processing stage.
This design offloads symbolic overhead and pointer tracking from the model's context window.
In addition, we encode high-level multi-agent behaviors such as queuing and group formation as structured BT subtrees.
This design decouples high-level intent from geometric execution while increasing behavioral diversity in the node database.
We further introduce a new node, \texttt{FollowVelocityField}, to support scalable group-level motion control.
Each group is associated with an independent velocity field that governs collective flow while the BT maintains decision-level control.
This hybrid design enables scalable crowd control through velocity fields while preserving structured decision logic through behavior trees.
To our knowledge, this is one of the first integrations of velocity-field crowd control directly into a behavior tree-based language-conditioned generation framework.

\paragraph{RAG module}
The BT XML schema contains heterogeneous node types with strict parameter constraints and non-trivial semantic dependencies.
Providing the full node specification inside the LLM prompt introduces excessive context length and amplifies hallucination risk.
We therefore design a RAG module that dynamically injects only task-relevant node specifications.
The retrieval corpus encodes node descriptions, parameter types, semantic constraints, and template structures.
Given a user prompt, the retriever selects the most relevant node specifications from the database.
These specifications are injected into the LLM context to guide schema generation.
This mechanism improves syntactic validity, reduces hallucinated node types, and accelerates inference by minimizing unnecessary context.

\subsubsection{Global planner on behavior trees}
While the SFM handles local collision avoidance among agents, it does not guarantee collision-free motion with respect to obstacles in densely structured environments as there are no hard constraints in SFMs to avoid situations where forces cancel out each other, leading to collision situations.
A geometric global planner is therefore integrated to ensure static obstacle avoidance.

\paragraph{Static Obstacle Avoidance via Theta*}
Thanks to our semantic world representation, our world can easily be decomposed into a 2D occupancy grid. 
This grid map enables the use of classical graph-based planners.
We adopt Theta* to compute efficient and near-line-of-sight collision-free paths between goal locations.
Theta* is chosen for its ability to generate shorter and smoother paths compared to standard A*~\cite{hart1968formal} while maintaining computational efficiency.

\paragraph{Planning Strategy}
Direct navigation command nodes (i.e. \texttt{GoTo} node) within the behavior tree are analyzed to identify consecutive goal-directed actions.
For each adjacent pair of navigation goals, a global path is computed on the occupancy grid.
The resulting intermediate waypoints are injected into the behavior tree as explicit sub-goals.
By performing planning at the behavior tree level, we preserve hierarchical structure while guaranteeing static obstacle avoidance.
This integration establishes a principled connection between symbolic task programs and classical motion planning.
}

\subsection{User Interface}

{

A GUI within RViz2 allows users to input prompts, select presets, and verify LLM-generated velocity fields and waypoints before committing to full physical simulation.
\autoref{img:ui} showcases the workflow 

}

\subsection{Optimized Presets}

{
To further reduce generation latency and improve controllability for common scenario types, we introduce \emph{presets} as structured constraint layers applied to the behavior generation and planning stages.
Each preset modifies specific components of the pipeline, including the LLM prompt template, retrievable node set, goal sampling strategy, velocity control policy, and global planning configuration.
Rather than altering the architecture itself, presets operate as deterministic configuration overlays.

\subsubsection{Emergency Mode}
Emergency Mode targets evacuation scenarios characterized by high-density crowds, urgency-driven motion, and dominant exit-seeking behavior.
The LLM prompt template is biased toward evacuation semantics, and the RAG module restricts retrieval to \texttt{FollowVelocityField}-based navigation to ensure scalability and coherent group-level motion.
Complex social interaction nodes such as conversation, roaming, or discretionary following are disabled to eliminate non-essential branching.
In this preset, the LLM generates spawn positions, group assignments for each pedestrian, and a single exit door represented as a 2D point $e$ shared by all pedestrians. 
For each group $\mathcal{G}_i$, we compute its centroid $c_i$ using $c_i = \frac{1}{|\mathcal{G}i|}\sum_{p\in\mathcal{G}_i} p$, where $p$ denotes an individual pedestrian position.

Theta* computes a feasible path between each group centroid $c_i$ and the exit $e$. 
These paths are subsequently converted into a velocity field, following the procedure in~\cite{textcrowd}, to generate supervision signals analogous to the label velocity fields used for training their diffusion-based velocity field model. 
This yields more stable and globally consistent guidance, while overcoming the limitation of a fixed set of predefined start and goal regions.
Overall, this configuration reformulates the generation task from compositional social reasoning to constrained crowd flow optimization.

\subsubsection{Queuing Mode}
Queuing Mode targets structured waiting scenarios such as service counters, ticket booths, or checkpoints where ordered spatial arrangement is required.
The behavior generation stage activates a dedicated \emph{Queue} subtree template that encodes ordered agent insertion and spacing constraints within a structured behavior tree.
Goal assignment follows a sequential allocation policy consistent with queue ordering rather than independent sampling.
In contrast to Emergency Mode, the Theta*-based global planner remains enabled to guarantee static obstacle avoidance in confined service areas.
The RAG module prioritizes queue-related node definitions while preserving geometric feasibility through global planning.

\subsubsection{Normal Mode}
Normal Mode represents daily-life mixed scenarios involving heterogeneous social interactions and moderate crowd density.
The full social behavior node set is available, including group formation, conversation, following, and general roaming behaviors.
Both hierarchical behavior tree control and Theta*-based global planning are enabled to balance expressiveness and geometric feasibility.
This mode preserves maximal behavioral diversity while maintaining hierarchical feasibility guarantees across all layers.

\begin{figure*}[t]
    \includegraphics[width=.99\linewidth]{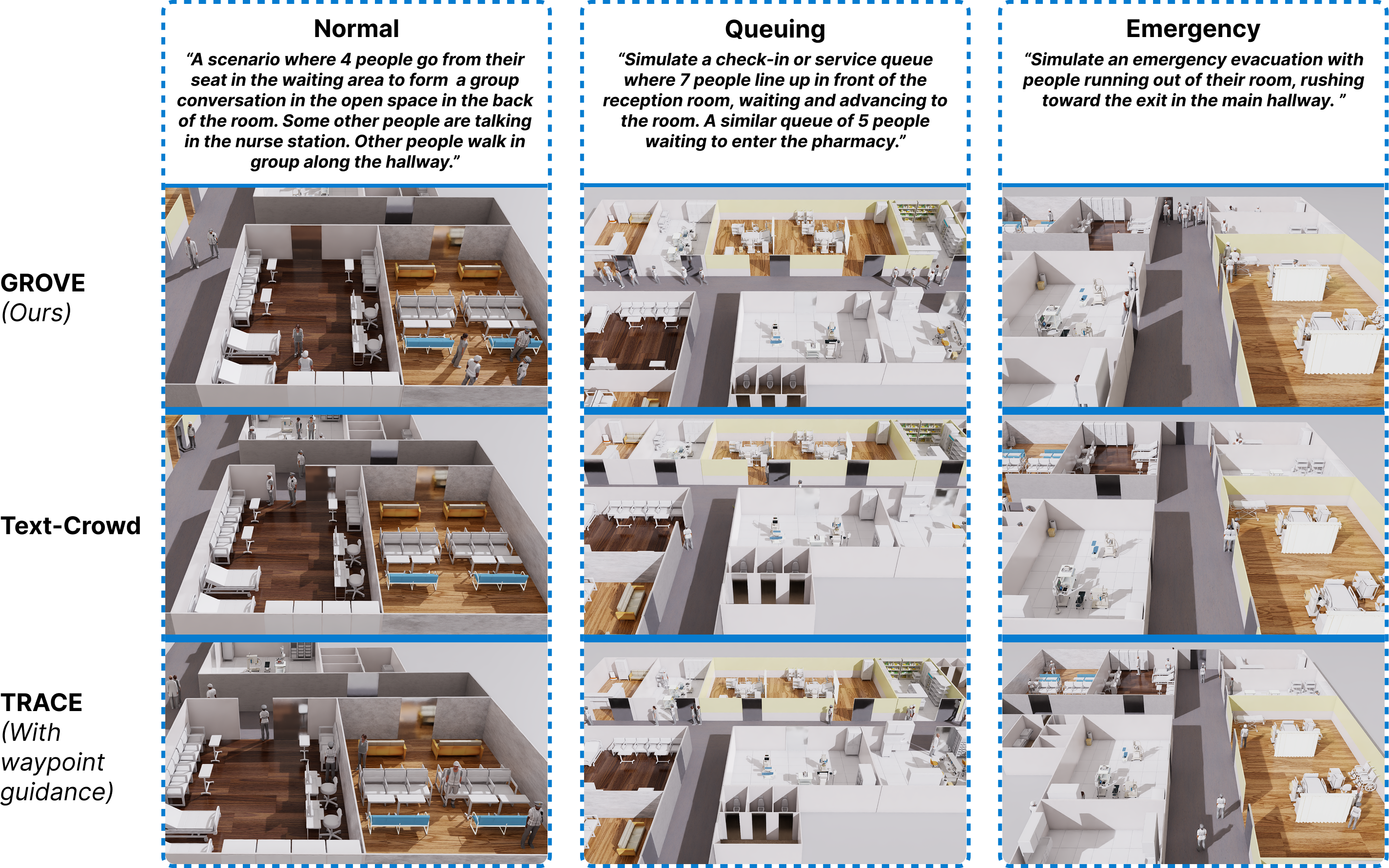}
    \caption{
    Side-by-side scenario comparison under identical prompts across Normal, Queuing, and Emergency presets in the Hospital world. \WORKNAME produces semantically coherent spatial organization — notably structured queue formation and exit-oriented crowd flow — that is absent or weakly represented in both baselines.
    }
    \label{fig:qualitative_comparison}
\end{figure*}


}

%% file: content/4-evaluation.tex
{
The proposed hierarchical pipeline is evaluated against two existing language-driven crowd generation frameworks across three scenario presets. 
We evaluate semantic alignment and realism quantitatively via VLM-based scoring
and navigation behavior through trajectory visualization across three scenario presets.
Generalization is tested across structured public, semi-structured professional, and unstructured domestic layouts through experiments in hospital, office, and residential environments.
The hospital environment, offering the richest semantic structure and most challenging navigation layout among the three, serves as our primary evaluation setting.
The comparison baselines are our re-implementations of Text-Crowd \cite{textcrowd} and TRACE \cite{tracepace}, which represent two recent crowd generation paradigms with differing assumptions on world representation and agent control.
To ensure comparability despite architectural differences, we adapt each environment to satisfy the input requirements of the baselines while preserving the underlying geometry and semantic structure.
Text-Crowd's limited semantic map format was derived automatically from \texttt{world.yaml} by directly mapping our semantic annotations.
TRACE provides neither a spawning sampler, nor an explicit waypoint generation mechanism.
LLM-generated spawn positions and a single waypoint per agent were provided as minimal guidance for the model to function properly.
This setup ensures that both baselines operate under comparable semantic input and geometric constraints without artificially enhancing their capabilities.

Preliminary experiments confirmed that providing the full node specification without retrieval consistently increased hallucinated node types and inference latency; the RAG module was adopted as a direct result of these observations.
Similarly, omitting the global planner produced frequent wall collisions and semantically implausible straight-line paths in structured environments, motivating its integration as a hard geometric feasibility guarantee.

}

\begin{figure*}[t]
    \centering
    \begin{subfigure}[t]{0.32\linewidth}
        \centering
        \includegraphics[width=\linewidth]{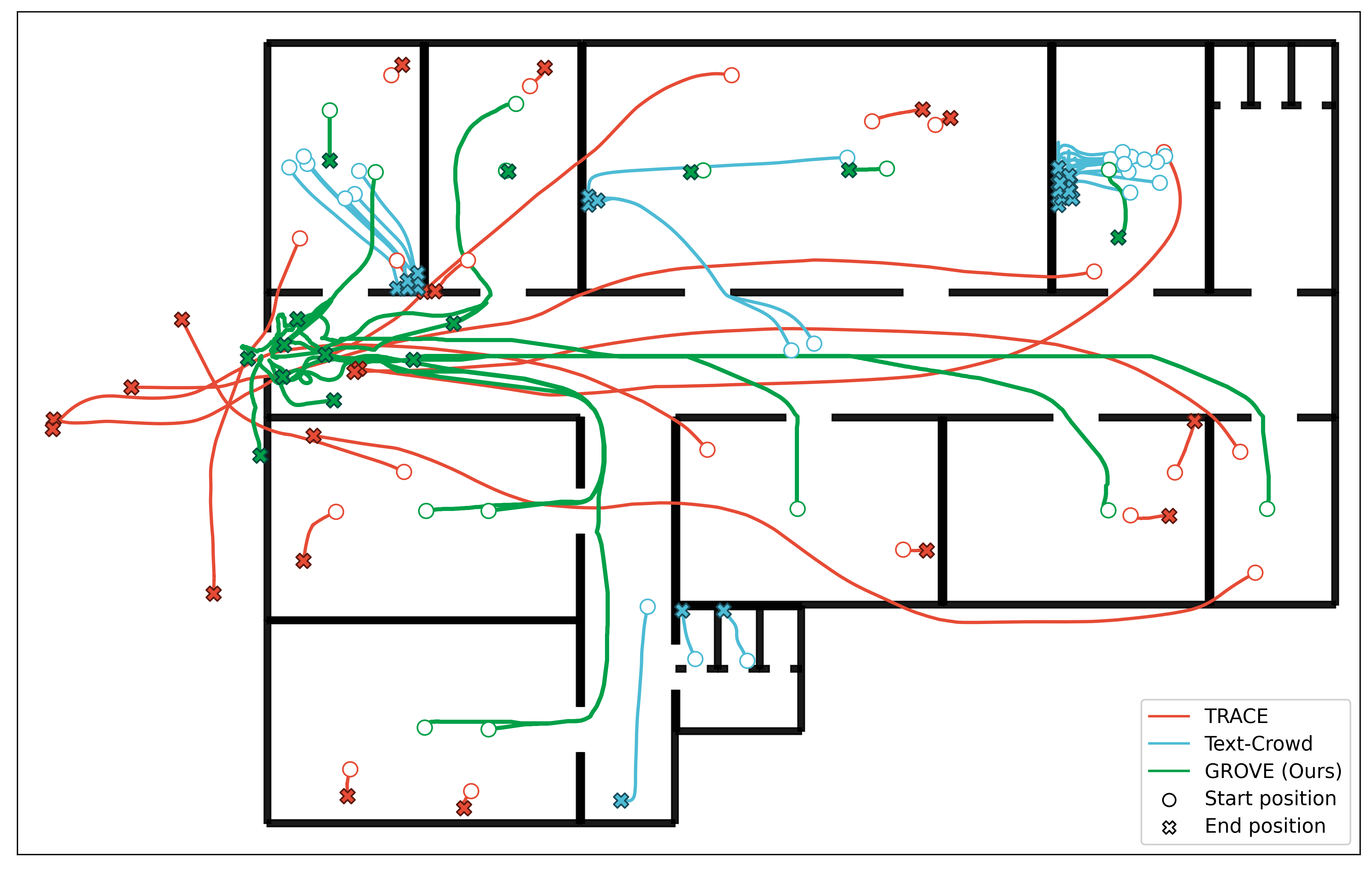}
        \caption{Scenario Emergency}
        \label{fig:scenario_emergency}
    \end{subfigure}
    \hfill
    \begin{subfigure}[t]{0.32\linewidth}
        \centering
        \includegraphics[width=\linewidth]{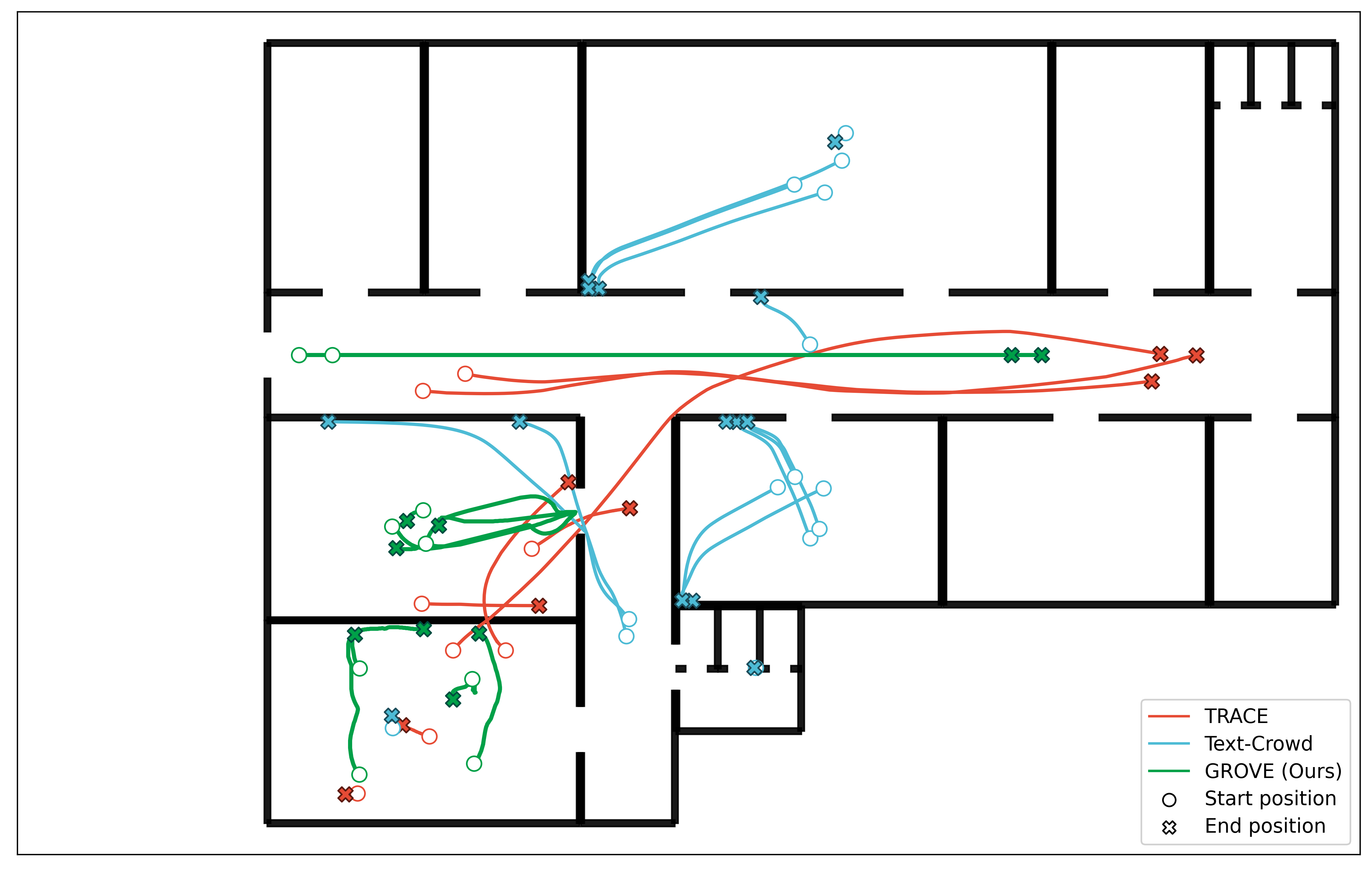}
        \caption{Scenario Normal}
        \label{fig:scenario_normal}
    \end{subfigure}
    \hfill
    \begin{subfigure}[t]{0.32\linewidth}
        \centering
        \includegraphics[width=\linewidth]{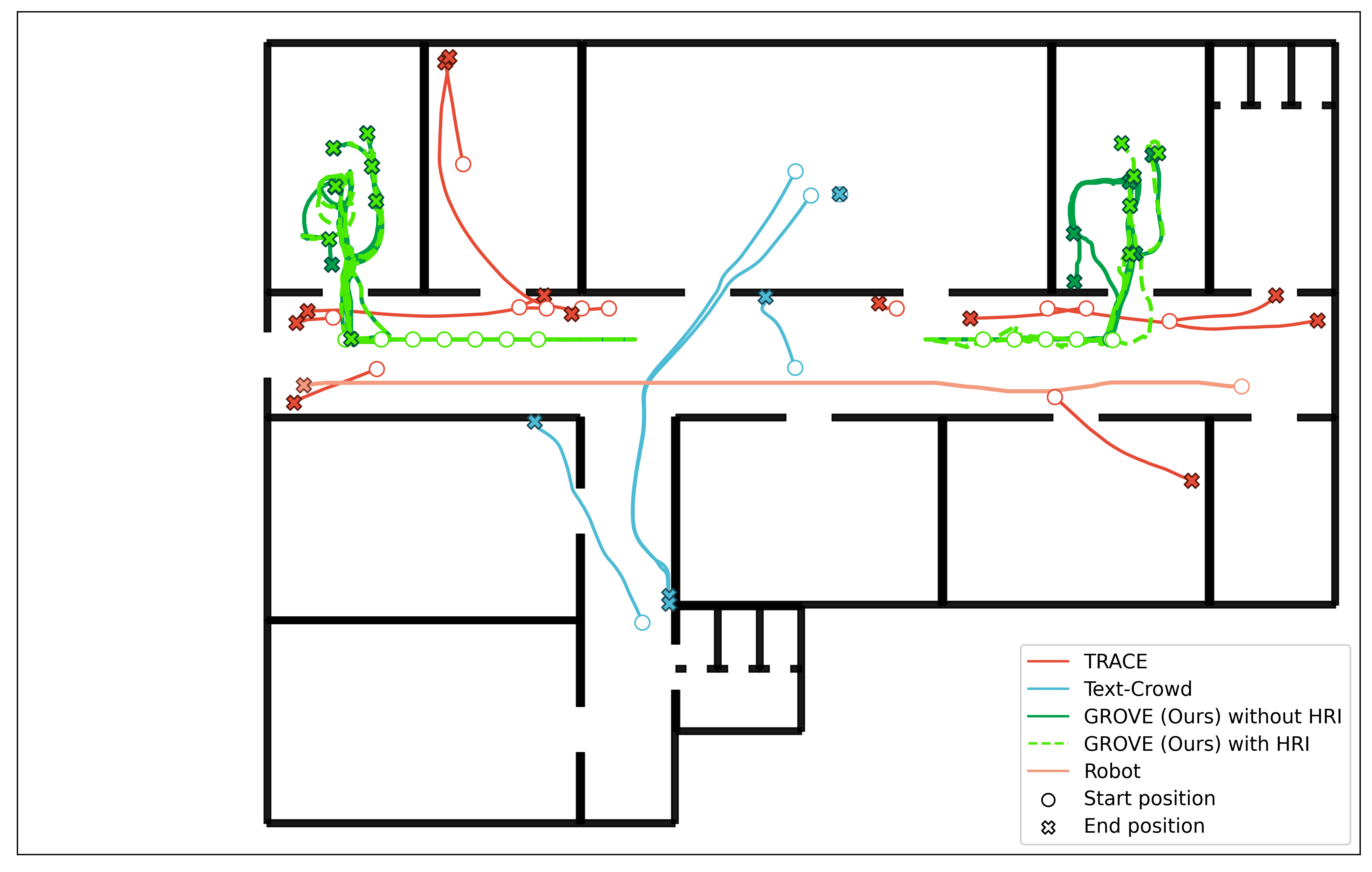}
        \caption{Scenario Queuing}
        \label{fig:scenario_queuing}
    \end{subfigure}
    
    \caption{Trajectory generation comparison across TRACE, Text-Crowd, and \WORKNAME in the Hospital world. (a) Emergency: \WORKNAME's velocity-field guidance produces coherent exit-directed flow without wall penetration, unlike TRACE which shows frequent wall collisions and incomplete exit-seeking. (b) Normal: \WORKNAME exhibits long-horizon goal consistency and stop-and-go patterns absent in both baselines. (c) Queuing: \WORKNAME is the only method producing structured queues with progressive agent advancement; note agents near the pharmacy adjusting toward the robot while maintaining queue structure.}
    \label{fig:traj_comparison}
\end{figure*}

\subsection{Path Generation}
{
\autoref{fig:traj_comparison} shows representative trajectory results focuses on static obstacle avoidance, semantic alignment with the prompt, and long-horizon navigation consistency.

\paragraph{TRACE}
TRACE produces visually smooth short-horizon trajectories in all scenarios.
However, in the Emergency scenario in \autoref{fig:scenario_emergency}, although the LLM assigns goals near the exit, only a subset of agents consistently move toward it even with waypoint guidance enabled.
We observe frequent wall penetration across scenarios, which we attribute to thin-wall structures that are weakly encoded in its agent-centric top-down semantic representation.

\paragraph{Text-Crowd}
Text-Crowd demonstrates stronger static obstacle avoidance due to ORCA-based post-processing that enforces collision-free velocity updates.
Nevertheless, trajectory endpoints are often weakly aligned with prompt semantics.
Although candidate start and goal regions are automatically derived from zone centers to reduce manual bias, goal sampling remains decoupled from fine-grained semantic reasoning.
Agents therefore terminate in geometrically valid but semantically implausible positions.

\paragraph{\WORKNAME}
Our method maintains collision-free trajectories while preserving strong semantic grounding across all presets.
In Emergency Mode, velocity-field guidance produces coherent crowd flow toward exits without wall penetration.
In Queuing and Normal modes, behavior tree execution preserves long-horizon intent, while Theta*-based waypoint injection guarantees geometric feasibility.
Compared to TRACE and Text-Crowd, goal allocation is directly grounded in RoI extraction, preventing semantically implausible terminal states.
Notably, \WORKNAME is the only method capable of generating structured queues in which agents line up and progressively advance toward the referenced region.
Furthermore, the use of behavior trees enables explicit stop-and-go patterns that frequently occur in real-world pedestrian dynamics but are absent in the baselines.
As shown in \autoref{fig:scenario_queuing}, agents near the pharmacy queue adjust their motion toward the robot while maintaining queue structure, producing realistic interactive behavior.
These results demonstrate that coupling symbolic task structure with geometric certification improves semantic fidelity, structural realism, and navigation robustness.
}

\subsection{Realism and Complexity}

{
Evaluations relying on 
trajectory-based metrics cannot capture alignment between prompt semantics and global scene structure.
We therefore adopt a Vision-Language Model (VLM)-based evaluation protocol following~\cite{tam2026sceneeval}, which demonstrated VLM-based scoring correlates strongly with human judgment in scene evaluation tasks.
For each method and scenario, we collect multiple screenshots from different viewpoints together with the annotated floor plan containing zone names.
These images, along with the original user prompt, are evaluated with \texttt{GPT-5}, a state-of-the-art VLM with strong vision-language reasoning capabilities.
The model is instructed to rate each scenario along three dimensions, namely
\emph{Prompt Alignment} (correspondence between the generated scene and the user’s semantic intent),
\emph{Plausibility} (could the scenario reasonably occur in the real world without counterintuitive artifacts?),
\emph{Visual Realism} (perceptual credibility of pedestrian appearance and spatial arrangement)\\
on a 0–10 scale.

\begin{table}[]
    \centering
    \begin{tabular}{lcccc}
    \toprule
    \textbf{Model} & \textbf{Alignment} & \textbf{Plausibility} & \textbf{Visual} & \textbf{Average} \\
    \midrule
    \multicolumn{5}{c}{\textbf{Emergency}} \\
    TRACE & 2.09 & 5.55 & 4.64 & 4.09 \\
    Text-Crowd & 2.71 & 5.71 & \textbf{4.86} & 4.43 \\
    GROVE & \textbf{4.50} & \textbf{6.67} & 4.83 & \textbf{5.33} \\
    \midrule
    \multicolumn{5}{c}{\textbf{Normal}} \\
    Text-Crowd & 2.50 & 6.67 & 5.50 & 4.89 \\
    TRACE & 4.17 & 7.17 & 5.17 & 5.50 \\
    GROVE & \textbf{6.29} & \textbf{7.71} & \textbf{5.71} & \textbf{6.57} \\
    \midrule
    \multicolumn{5}{c}{\textbf{Queuing}} \\
    TRACE & 2.83 & 5.83 & 4.50 & 4.39 \\
    Text-Crowd & 1.50 & 7.17 & 5.17 & 4.61 \\
    GROVE & \textbf{6.86} & \textbf{7.43} & \textbf{5.57} & \textbf{6.62} \\
    \bottomrule
    \end{tabular}
    \caption{Quantitative Evaluations of Scenarios (0–10 scale, $N=3$ scenarios per preset, $K=64$ screenshots total). \WORKNAME achieves the highest average across all presets, with the largest gains in Prompt Alignment.}
    \label{tab:vlm_eval}
\end{table}

Results are reported in \autoref{tab:vlm_eval}.
Across all presets, \WORKNAME achieves the highest average score.
The most significant improvement is observed in Prompt Alignment, where \WORKNAME consistently outperforms both baselines.
This reflects the effectiveness of RoI-grounded start and goal allocation.
Text-Crowd exhibits comparatively low alignment scores, particularly in Emergency and Queuing scenarios, due to its decoupled start and goal sampling strategy.
TRACE achieves moderate plausibility and visual realism scores but lower alignment, indicating smooth local motion without strong semantic control.
Notably, in the Queuing preset, \WORKNAME substantially improves both alignment and plausibility, demonstrating its ability to generate structured multi-agent behaviors that are absent or weakly represented in the baselines.
A side-by-side qualitative comparison in \autoref{fig:qualitative_comparison} corroborates these findings, where \WORKNAME produces semantically coherent spatial organization and goal-consistent agent configurations.
The qualitative samples further reveal clearer queue formation, exit-oriented crowd flow, and interaction-consistent grouping patterns.
Together, these quantitative and visual results indicate that coupling semantic grounding with hierarchical planning improves geometric correctness, structural organization, and perceived scenario realism.
}

\subsection{Optimized Presets}
{
We evaluate the effectiveness of our optimized presets against a vanilla variant of \WORKNAME. 
In the vanilla setting, the full world description is injected into the prompt, no restriction is imposed on the RAG candidate behavior nodes, and the model autonomously selects behavior nodes without scenario-specific system guidance.
Results are reported in \autoref{tab:token_count}.

\begin{table}[h]
    \centering
    \begin{tabular}{lccc}
    \toprule
    \textbf{Method} & \textbf{Emergency} & \textbf{Normal} & \textbf{Queuing} \\
    \midrule
    Vanilla & 14296 & 14329 & 14343 \\
    \WORKNAME & 9250 & 7323 & 8575 \\
    \midrule
    Improvement & $35\%$ & $49\%$ & $40\%$ \\
    \bottomrule
    \end{tabular}
    \caption{Token count, averaged over $N=10$ cold-cache runs (lower is better). \WORKNAME presets provide significant double digit token efficiency improvements.}
    \label{tab:token_count}
\end{table}





%% file: content/5-conclusion.tex
In this paper, we introduced \WORKNAME as a novel approach to pedestrian simulation, using generative models to combine multiple SotA approaches into an adaptive, high-fidelity simulation.
We have validated our system against established baselines and demonstrated a more accessible, efficient, and more behaviorally realistic pedestrian simulation directly compatible with ROS2 and popular robot simulators.

Several limitations remain.
The computational cost of combining multiple state-of-the-art approaches is high; while we streamlined the generation process and simulation loop where possible, inference remains expensive compared to most existing approaches.
Additionally, our pipeline assumes a static obstacle map and does not account for dynamic environmental changes such as moving furniture or temporary obstructions.
We aim to address the computational cost in future work by introducing world-level caches pre-computed ahead of time, and plan to explore automated world representation extraction from sensor data to reduce manual authoring overhead.

%% file: references.bib
@inproceedings{arena:5,
	AUTHOR = {Linh Kästner AND Volodymyr Shcherbyna AND Harold Soh AND Giang Nguyen Huu Truong AND Do Duc Anh AND Ton Manh Kien AND Tim Seeger AND Ahmed Martban AND Vu Thanh Lam AND Nguyen Quoc Hung AND Pham Thai Hoang Tung AND Tran Dang An AND Eva Wiese AND Maximilian Ho-Kyoung Schreff},
	TITLE = {{Demonstrating Arena 5.0: A Photorealistic ROS2 Simulation Framework for Developing and Benchmarking Social Navigation}},
	BOOKTITLE = {Proceedings of Robotics: Science and Systems},
	YEAR = {2025},
	ADDRESS = {LosAngeles, CA, USA},
	MONTH = {June},
	DOI = {10.15607/RSS.2025.XXI.092}
}

@misc{hunavsim2,
  author        = {Escudero, Iv{\'a}n and P{\'e}rez-Higueras, No{\'e} and Caballero, Fernando and Merino, Luis},
  title         = {{HuNavSim}~2.0: An Extended {ROS}~2 Human Navigation Simulator},
  year          = {2025},
  eprint        = {2507.17317},
  archivePrefix = {arXiv},
  primaryClass  = {cs.RO},
  url           = {https://arxiv.org/abs/2507.17317}
}

@article{socnavbench,
  author  = {Biswas, Abhijat and Wang, Allan and Silvera, Gustavo and Steinfeld, Aaron and Admoni, Henny},
  title   = {{SocNavBench}: A Grounded Simulation Testing Framework for Evaluating Social Navigation},
  journal = {ACM Transactions on Human-Robot Interaction},
  year    = {2022},
  volume  = {11},
  number  = {3},
  pages   = {26:1--26:24},
  doi     = {10.1145/3476413}
}

@article{sean,
  author    = {Tsoi, Nathan and Xiang, Alec and Yu, Peter and Sohn, Samuel S. and Schwartz, Greg and Ramesh, Subashri and Hussein, Mohamed and Gupta, Anjali W. and Kapadia, Mubbasir and V{\'a}zquez, Marynel},
  journal   = {IEEE Robotics and Automation Letters},
  title     = {{SEAN}~2.0: Formalizing and Generating Social Situations for Robot Navigation},
  year      = {2022},
  volume    = {7},
  number    = {4},
  pages     = {11047--11054},
  doi       = {10.1109/LRA.2022.3196783}
}

@inproceedings{socialgym,
  author    = {Holtz, Jarrett and Biswas, Joydeep},
  title     = {{SOCIALGYM}~2.0: Simulator for Multi-Robot Learning and Navigation in Shared Human Spaces},
  booktitle = {Proceedings of the AAAI Conference on Artificial Intelligence},
  year      = {2024},
  volume    = {38},
  number    = {21},
  pages     = {23778--23780},
  doi       = {10.1609/aaai.v38i21.30562}
}

@inproceedings{textcrowd,
  author    = {Ji, Xuebo and Pan, Zherong and Gao, Xifeng and Pan, Jia},
  title     = {Text-Guided Synthesis of Crowd Animation},
  booktitle = {ACM SIGGRAPH 2024 Conference Papers},
  year      = {2024},
  doi       = {10.1145/3641519.3657516}
}

@inproceedings{tracepace,
  author    = {Rempe, Davis and Luo, Zhengyi and Peng, Xue Bin and Yuan, Ye and Kitani, Kris and Kreis, Karsten and Fidler, Sanja and Litany, Or},
  title     = {Trace and Pace: Controllable Pedestrian Animation via Guided Trajectory Diffusion},
  booktitle = {Proceedings of the IEEE/CVF Conference on Computer Vision and Pattern Recognition (CVPR)},
  year      = {2023},
  pages     = {13756--13766}
}

@article{helbing1995sfm,
  author  = {Helbing, Dirk and Moln{\'a}r, P{\'e}ter},
  title   = {Social Force Model for Pedestrian Dynamics},
  journal = {Physical Review E},
  year    = {1995},
  volume  = {51},
  number  = {5},
  pages   = {4282--4286},
  doi     = {10.1103/PhysRevE.51.4282}
}

@article{moussaid2010walking,
  title={The walking behaviour of pedestrian social groups and its impact on crowd dynamics},
  author={Moussa{\"\i}d, Mehdi and Perozo, Niriaska and Garnier, Simon and Helbing, Dirk and Theraulaz, Guy},
  journal={PloS one},
  volume={5},
  number={4},
  pages={e10047},
  year={2010},
  publisher={Public Library of Science San Francisco, USA}
}

@article{rudenko2020survey,
  title     = {Human Motion Trajectory Prediction: A Survey},
  author    = {Rudenko, Andrey and Palmieri, Luigi and Herman, Michael and Kitani, Kris M. and Gavrila, Dariu M. and Arras, Kai O.},
  journal   = {The International Journal of Robotics Research},
  volume    = {39},
  number    = {8},
  pages     = {895--935},
  year      = {2020},
  publisher = {SAGE},
  doi       = {10.1177/0278364920917446}
}

@article{mavrogiannis2023eval,
  title     = {Evaluation of Socially-Aware Robot Navigation},
  author    = {Mavrogiannis, Christoforos and Baldini, Francesca and Wang, Allan and Zhao, Dapeng and Trautman, Pete and Steinfeld, Aaron and Oh, Jean},
  journal   = {Frontiers in Robotics and AI},
  volume    = {8},
  year      = {2021},
  publisher = {Frontiers},
  doi       = {10.3389/frobt.2021.721317}
}

@article{chen2018sfmreview,
  title     = {Social Force Models for Pedestrian Traffic -- State of the Art},
  author    = {Chen, Xu and Treiber, Martin and Kanagaraj, Venkatesan and Li, Haiying},
  journal   = {Transport Reviews},
  volume    = {38},
  number    = {5},
  pages     = {625--653},
  year      = {2018},
  publisher = {Taylor \& Francis},
  doi       = {10.1080/01441647.2017.1396265}
}

@inproceedings{zhao2020sim2real,
  title     = {Sim-to-Real Transfer in Deep Reinforcement Learning for Robotics: A Survey},
  author    = {Zhao, Wenshuai and Queralta, Jorge Pe{\~n}a and Westerlund, Tomi},
  booktitle = {IEEE Symposium Series on Computational Intelligence (SSCI)},
  pages     = {737--744},
  year      = {2020},
  doi       = {10.1109/SSCI47803.2020.9308468}
}

@inproceedings{wang2023gensim,
  title     = {{GenSim}: Generating Robotic Simulation Tasks via Large Language Models},
  author    = {Wang, Lirui and Ling, Yiyang and Yuan, Zhecheng and Shridhar, Mohit and Bao, Chen and Qin, Yuzhe and Wang, Bailin and Xu, Huazhe and Wang, Xiaolong},
  booktitle = {International Conference on Learning Representations (ICLR)},
  year      = {2024}
}

@inproceedings{tam2026sceneeval,
  title={SceneEval: Evaluating semantic coherence in text-conditioned 3D indoor scene synthesis},
  author={Tam, Hou In Ivan and Pun, Hou In Derek and Wang, Austin T and Chang, Angel X and Savva, Manolis},
  booktitle={Proceedings of the IEEE/CVF Winter Conference on Applications of Computer Vision},
  pages={7355--7365},
  year={2026}
}

@article{daniel2010theta,
  title={Theta*: Any-angle path planning on grids},
  author={Daniel, Kenny and Nash, Alex and Koenig, Sven and Felner, Ariel},
  journal={Journal of Artificial Intelligence Research},
  volume={39},
  pages={533--579},
  year={2010}
}

@inproceedings{li2025learning,
  title={Learning better representations for crowded pedestrians in offboard lidar-camera 3d tracking-by-detection},
  author={Li, Shichao and Li, Peiliang and Lian, Qing and Yun, Peng and Chen, Xiaozhi},
  booktitle={2025 IEEE International Conference on Robotics and Automation (ICRA)},
  pages={2740--2747},
  year={2025},
  organization={IEEE}
}

@inproceedings{jiang2025physgcn,
  title={PhysGCN-DL: Physics-Informed Graph Convolutional Networks with Diversity-Aware Loss Optimization for Multimodal Pedestrian Trajectory Prediction},
  author={Jiang, Zihan and Liu, Ruonan and Zhou, Yibo and Lu, Haibo and Yang, BoYuan and Lin, Di and Zhang, Weidong},
  booktitle={2025 IEEE/RSJ International Conference on Intelligent Robots and Systems (IROS)},
  pages={38--45},
  year={2025},
  organization={IEEE}
}

@inproceedings{shenkut2025visual,
  title={Visual-Linguistic Reasoning for Pedestrian Trajectory Prediction},
  author={Shenkut, Dereje and Kumar, BVK Vijaya},
  booktitle={2025 IEEE International Conference on Robotics and Automation (ICRA)},
  pages={771--778},
  year={2025},
  organization={IEEE}
}

@inproceedings{bokkasam2025pedestrian,
  title={Pedestrian Intention and Trajectory Prediction in Unstructured Traffic Using IDD-PeD},
  author={Bokkasam, Ruthvik and Gangisetty, Shankar and Hafez, AH Abdul and Jawahar, CV},
  booktitle={2025 IEEE International Conference on Robotics and Automation (ICRA)},
  pages={763--770},
  year={2025},
  organization={IEEE}
}

@inproceedings{chen2024social,
  title={Social physics informed diffusion model for crowd simulation},
  author={Chen, Hongyi and Ding, Jingtao and Li, Yong and Wang, Yue and Zhang, Xiao-Ping},
  booktitle={Proceedings of the AAAI Conference on Artificial Intelligence},
  volume={38},
  number={1},
  pages={474--482},
  year={2024}
}

@inproceedings{caesar2020nuscenes,
  title={nuscenes: A multimodal dataset for autonomous driving},
  author={Caesar, Holger and Bankiti, Varun and Lang, Alex H and Vora, Sourabh and Liong, Venice Erin and Xu, Qiang and Krishnan, Anush and Pan, Yu and Baldan, Giancarlo and Beijbom, Oscar},
  booktitle={Proceedings of the IEEE/CVF conference on computer vision and pattern recognition},
  pages={11621--11631},
  year={2020}
}

@article{comanici2025gemini,
  title={Gemini 2.5: Pushing the frontier with advanced reasoning, multimodality, long context, and next generation agentic capabilities},
  author={Comanici, Gheorghe and Bieber, Eric and Schaekermann, Mike and Pasupat, Ice and Sachdeva, Noveen and Dhillon, Inderjit and Blistein, Marcel and Ram, Ori and Zhang, Dan and Rosen, Evan and others},
  journal={arXiv preprint arXiv:2507.06261},
  year={2025}
}

@article{yang2025qwen3,
  title={Qwen3 technical report},
  author={Yang, An and Li, Anfeng and Yang, Baosong and Zhang, Beichen and Hui, Binyuan and Zheng, Bo and Yu, Bowen and Gao, Chang and Huang, Chengen and Lv, Chenxu and others},
  journal={arXiv preprint arXiv:2505.09388},
  year={2025}
}

@article{yao2020learning,
  title={Learning crowd behavior from real data: A residual network method for crowd simulation},
  author={Yao, Zhenzhen and Zhang, Guijuan and Lu, Dianjie and Liu, Hong},
  journal={Neurocomputing},
  volume={404},
  pages={173--185},
  year={2020},
  publisher={Elsevier}
}

@inproceedings{liu2024learning,
  title={Learning to generate diverse pedestrian movements from web videos with noisy labels},
  author={Liu, Zhizheng and Lin, Joe and Wu, Wayne and Zhou, Bolei},
  booktitle={The Thirteenth International Conference on Learning Representations},
  year={2024}
}

@inproceedings{bae2025continuous,
  title={Continuous locomotive crowd behavior generation},
  author={Bae, Inhwan and Lee, Junoh and Jeon, Hae-Gon},
  booktitle={Proceedings of the Computer Vision and Pattern Recognition Conference},
  pages={22416--22431},
  year={2025}
}

@inproceedings{shi2023learning,
  title={Learning to simulate crowd trajectories with graph networks},
  author={Shi, Hongzhi and Yao, Quanming and Li, Yong},
  booktitle={Proceedings of the ACM Web Conference 2023},
  pages={4200--4209},
  year={2023}
}

@article{li2025efficient,
  title={Efficient crowd simulation in complex environment using deep reinforcement learning},
  author={Li, Yihao and Chen, Yuting and Liu, Junyu and Huang, Tianyu},
  journal={Scientific Reports},
  volume={15},
  number={1},
  pages={5403},
  year={2025},
  publisher={Nature Publishing Group UK London}
}

@article{hart1968formal,
  title={A formal basis for the heuristic determination of minimum cost paths},
  author={Hart, Peter E and Nilsson, Nils J and Raphael, Bertram},
  journal={IEEE transactions on Systems Science and Cybernetics},
  volume={4},
  number={2},
  pages={100--107},
  year={1968},
  publisher={IEEE}
}
